\documentclass[conference]{IEEEtran}
\IEEEoverridecommandlockouts
\usepackage{cite}
\usepackage{amsmath,amssymb,amsfonts}
\usepackage{graphicx}
\usepackage{textcomp}
\usepackage{xcolor}
\usepackage{multirow}

\usepackage[utf8]{inputenc}
\usepackage{algorithm} 
\usepackage{algpseudocode}

\definecolor{darkgreen}{RGB}{0,120,0}

\DeclareMathOperator*{\argmax}{arg\,max}

\usepackage[a4paper, total={184mm,239mm}]{geometry}
\def\BibTeX{{\rm B\kern-.05em{\sc i\kern-.025em b}\kern-.08em
    T\kern-.1667em\lower.7ex\hbox{E}\kern-.125emX}}

\usepackage[super]{nth}

\definecolor{myc}{cmyk}{1,0,1,0}
\makeatletter
\renewcommand\footnoterule{%
  \kern-3\p@
  \hrule\@width.4\columnwidth
  \kern2.6\p@}
  \makeatother

\begin{document}

% dynamic neural architecture search - energy-efficient - DVFS - early exits - joint optimization - multi-branch - edge computing/edge-ai

\title{HADAS: \underline{H}ardware-\underline{A}ware \underline{D}ynamic Neural \underline{A}rchitecture Search for Edge Performance \underline{S}caling\\}

% title propositions
% #1: HADAS: Hardware-Aware Dynamic neural Architecture Search
% #2: HADS: Hardware-Aware Dynamic neural architecture Search
% #3: HW-dynamic-NAS: Hardware-Aware Dynamic Neural Architecture Search
% #4: Dynamic-Edge-NAS: Towards Efficient Dynamic Inference on Edge devices


% \author{\IEEEauthorblockN{}
% \IEEEauthorblockA{} \\
% \textit{}\\ 
% \textit{\{modema, alfaruqu\}}@uci.edu}\vspace{-8truemm}}

\author{
\IEEEauthorblockN{
Halima Bouzidi\IEEEauthorrefmark{1}\textsuperscript{\textsection},
Mohanad Odema\IEEEauthorrefmark{2}\textsuperscript{\textsection},
Hamza Ouarnoughi\IEEEauthorrefmark{1},
Mohammad Abdullah Al Faruque\IEEEauthorrefmark{2},
Smail Niar\IEEEauthorrefmark{1}}

\IEEEauthorrefmark{1}\textit{LAMIH/UMR CNRS, Université Polytechnique Hauts-de-France, Valenciennes, France} \\
\IEEEauthorrefmark{2}\textit{Department of Electrical Engineering and Computer Science, University of California, Irvine, USA} \\

\IEEEauthorrefmark{1}\textit{\{firstname.lastname\}}@uphf.fr
\hspace{20truemm}
\IEEEauthorrefmark{2}\textit{\{modema, alfaruqu\}}@uci.edu
\vspace{-5truemm}
}

\maketitle

\begingroup\renewcommand\thefootnote{\textsection}
\footnotetext{Equal contribution}
\endgroup

\begin{abstract}

% Dynamic neural networks (DyNNs) have become viable techniques to enable intelligence on resource-constrained edge devices while elevating computational efficiency. To incorporate model flexibility, existing design workflows entail morphing established neural architectures to integrate the dynamic computing components. Additionally, the heterogeneous edge computing components support hardware configuration features like dynamic voltage and frequency scaling (DVFS) for resource efficiency. To avert sub-optimal DyNN designs at the edge, we present a novel Dynamic Neural Architecture Search framework -- \textsc{HADAS} -- that optimizes the design process of DyNNs through the joint optimization of the entire DyNN componenets with the hardware configuration parameters. \textsc{HADAS} involves a bi-stage optimization is compatible with existing state-of-the-art  Considering these points, sub-optimal DyNN designs can be obtained as: (\emph{i}) the backbone architectures were originally optimized for \textit{rigid} model deployment, and (\emph{ii}) the negligence of the inter-dependencies between the design spaces of the DyNN architecture and hardware configuration parameters. 
% This paper presents  

Dynamic neural networks (DyNNs) have become viable techniques to enable intelligence on resource-constrained edge devices while maintaining computational efficiency. 
In many cases, the implementation of DyNNs can be sub-optimal due to its underlying backbone architecture being developed at the design stage \textit{independent} of both: (\emph{i}) the dynamic computing features, e.g. early exiting, and (\emph{ii}) the resource efficiency features of the underlying hardware, e.g., dynamic voltage and frequency scaling (DVFS). 
%(i) backbones were originally developed for rigid model deployment, and (ii) the negligence of the inter-dependencies between the design spaces of the DyNN architecture and hardware configuration parameters.
Addressing this, we present HADAS, a novel Hardware-Aware Dynamic Neural Architecture Search framework that realizes DyNN architectures whose backbone, early exiting features, and DVFS settings have been \textit{jointly} optimized to maximize performance and resource efficiency. 
Our experiments using the CIFAR-100 dataset and a diverse set of edge computing platforms have seen HADAS dynamic models achieve up to 57\% energy efficiency gains compared to the conventional dynamic ones while maintaining the desired level of accuracy scores. 
Our code is available at https://github.com/HalimaBouzidi/HADAS

\end{abstract}

% \vspace{-3ex}

\section{Introduction} 
\label{sec:introduction}

% problems exist when it comes to from a designer's perspective (monolithic rigid models as well) -- mention some examples on branchynet. design automation frameworks target single platform deployment

% Not to mention the dynamicity of the hardware is often also overlooked during design, and such variation can lead to suboptimal solutions 

% some works proposed to have multiple architectures on the edge device but that approach is challenging because 1,2,3

% what are the research challenges or where is the lacking in current SOTA

% what do we do to solve them 

%(\blue{Our edge over dynamic OFA paper is that we do not need to have an ensemble of models for different constraints, instead leverage branching from the same backbone for different purposes}

%early classifiers are not able to leverage the semantic-level features produced by the deeper layers, potentially causing significant accuracy drops. Hence, approaches like MSDNet (complex and needs specialized design) and S2 offer more sophisticated training/design techniques.) but still the generality of the approach is attractive in the sense that it is not specialized, but entails an additonal simple step to incorporate within the design framework, generalizable to other kinds of architectures such as transformers maybe
% \textcolor{purple}{It is Smail. I put my suggestions in purple. General remark: Some sentences are very long. I suggest to split them in two or 3.}
%%% DyNNs
Neural Networks (NNs) have become integral machine learning techniques that enable intelligence for today's edge computing applications. 
Oftentimes, edge computing platforms are deployed in-the-wild, making them susceptible to considerable runtime variations related to the distribution of collected data, i.e., difficulty of accurately processing an input, and  the system state, e.g., state of charge. 
Accordingly, the adoption of Dynamic Neural Networks (DyNNs) \cite{han2021dynamic} has become increasingly relevant, where in opposition to the conventional \textit{static} models with fixed computational graphs, DyNNs adapt their model structure or parameters to suit the corresponding runtime context.  
Consequently, DyNNs can offer resource efficiency gains at the edge while maintaining the models' utility.

%%% DyNNS + Early exit
One prominent DyNN technique is early exiting, where dynamic depth variation is applied on a sample-wise basis to avoid redundant computations. 
Specifically, early-exiting facilitates concluding the processing of the ``easier'' input samples at earlier layers of a model for resource efficiency. 
This feature is often realized through a multi-exit architecture that integrates intermediate classifiers onto a shared backbone model \cite{teerapittayanon2016branchynet, odema2021eexnas, panda2016conditional}. 

%  only the minimum number of computing components of a model along the execution path are activated during inference, that is, canonical (simple) inputs do not cause a full invocation of the entire model components to be processed effectively. This not only promotes computational efficiency, but also preserves the model's representational power through invoking the full model when harder data samples are encountered. Most notably, early-exiting has become a renowned technique to achieve that as the processing of an input sample can be concluded at the earlier shallow layers of a model whenever the input sample is deemed simple enough. In practice, early-exiting is often achieved through a multi-exit architecture that integrates intermediate classifiers onto a single shared backbone network to avoid redundant computations, with the input-to-exit mapping decision typically being made based on confidence estimates and/or predictive controllers \cite{teerapittayanon2016branchynet, odema2021eexnas, panda2016conditional}. 

%%% DyNNs + NAS
Typically, the design workflow of multi-exit models initially assumes that the backbone's architecture has been \textit{optimally} designed to maximize performance on a target task. 
Evidently, backbones in related works were either based on renowned state-of-the-art NN architectures, e.g., ResNets in \cite{teerapittayanon2016branchynet}, or models rendered through the design automation frameworks of Neural Architecture Search (NAS) \cite{wang2020attentivenas}. 
This means that backbones were originally designed to serve as \textit{standalone} static models. 
Thus, a subject of debate is whether such design optimality of these models would hold when auxiliary tasks are added -- as in to serve as the backbone of a dynamic model.

% From here, we identify two key deficiencies that could lead to sub-optimal designs of the multi-exit: 1) The architectural design of the backbone model itself was not intended for adaptive/dynamic execution, and 2) The hardware settings of the edge computing platform can change during runtime. For the former, this is evident in the related works as early-exiting methods is integrated onto well known state-of-the-art (SOTA) static model architectures characterized by an element of \textit{rigidity}. For example in \cite{teerapittayanon2016branchynet}, the backbones were SOTA convolutional neural network architectures, e.g., ResNets, whose design has been optimized manually to maximize performance on specific tasks (e.g., image recognition) through static models. The same notion still persists today with the design automation frameworks for neural architecture search (NAS) \cite{wang2020attentivenas}, whose purpose is to effectively search through the architectural parameters design space to \textit{automatically} identify model architectures that can outperform their manually designed counterparts on the target tasks. Given how the dynamic characteristic behaviors of models are overlooked in these early design stages, a valid question here is whether these architectural design choices would remain ideal in the context of a collective multi-exit model given an auxiliary task of serving as the backbone of intermediate classifiers.

Even more so, the design stage of NN architectures usually entails treating the configurable hardware settings of the edge platforms as fixed constraints \cite{panda2016conditional, odema2021eexnas}, overlooking supported resource efficiency features such as dynamic voltage and frequency scaling (DVFS).  
Unfavorably, this may lead to inferior model designs as a result of disregarding the inter-dependencies between the \textit{model} and \textit{hardware} design spaces. 
Although recent works have attempted to remedy this deficiency for static NNs through joint optimization approaches \cite{lou2021dynamic}, addressing it for DyNNs is still highly understudied. 
In summary, the current state-of-the-art DyNN design workflow lacks in the following:
\begin{itemize}
    \item The backbone model architectures were not originally optimized for dynamic inference 
    \item The hardware configuration settings are treated as fixed constraints during the design process
    \item Modern NN design frameworks (e.g., NAS) do not characterize the runtime aspects of dynamic input mappings
\end{itemize}
\vspace{-2ex}
\begin{figure}[ht]
\centering
  \includegraphics[width=0.49\textwidth]{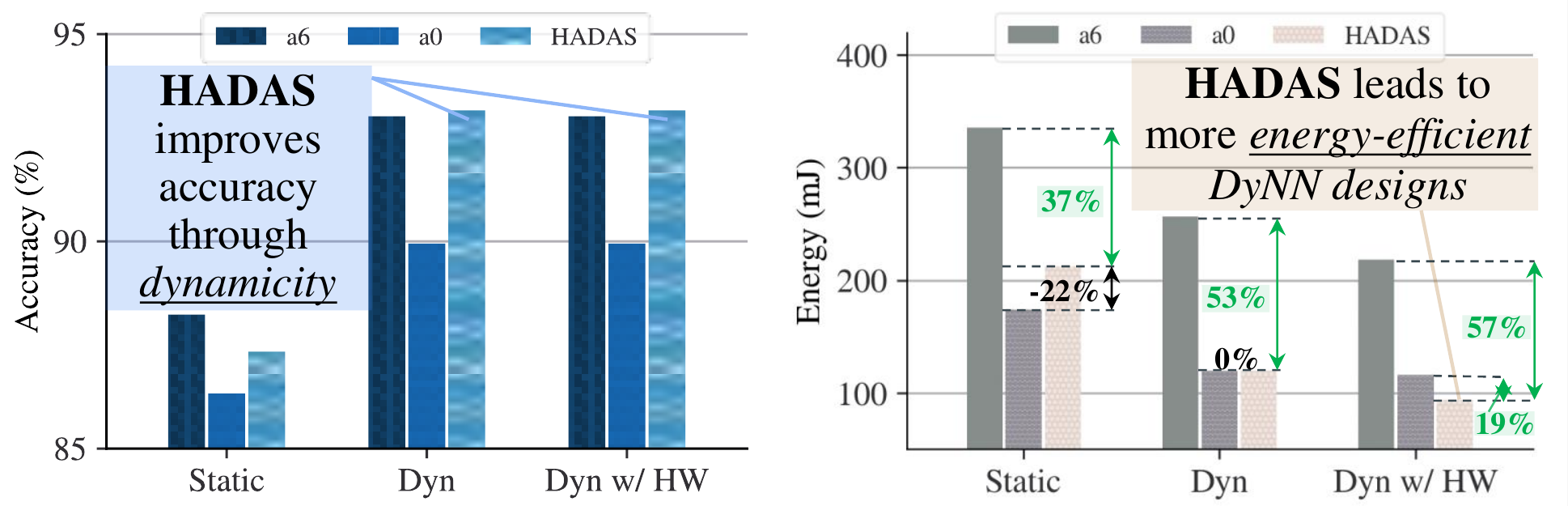}
  \vspace{-4.5ex}
  \caption{Comparing the performance of (a0, a6) from AttentiveNAS and HADAS's model on CIFAR-100 and the Jetson TX2 Pascal GPU hardware}
  \label{fig:motivation}
  \vspace{-2.2ex}
\end{figure}

\subsection{Motivational Example}
We take as baselines the respective \textit{most compact} and \textit{highest-performing} image recognition models, \textbf{a0} and \textbf{a6}, that were provided through a state-of-the-art NAS framework, AttentiveNAS \cite{wang2020attentivenas}. 
We compare their performance against one of our models that was provided using HADAS framework.
In this case, we implement HADAS on top of AttentiveNAS to ensure a fair comparison by having the models share the same base structure and optimization algorithms. 
Classification accuracy and energy consumption are leveraged as the performance comparison metrics. 
Here, we use the CIFAR-100 image dataset for models' training and accuracy evaluations and the NVIDIA Jetson TX2 platform for hardware benchmarking. 

As shown in Figure \ref{fig:motivation}, we designate three stages of optimizations that can be applied to maximize performance efficiency: \textbf{\textit{Static}} -- optimizing the backbone model design; \textbf{\textit{Dyn}} -- integrating dynamic \textit{early-exiting} features; and \textbf{\textit{Dyn w/ HW}} -- integrating early-exiting features and applying DVFS features. With regards to accuracy (\textit{left barplot}), HADAS's model outperforms \textbf{a0} and is on-par with \textbf{a6} after applying the \textit{static} and \textit{Dyn} optimizations. More interestingly, though, the energy efficiency of HADAS's model is enhanced considerably with every applied optimization compared to the other models (\textit{right barplot)}. 
After the first stage of \textbf{\textit{Static}} optimization, \textbf{a0} is reasonably deemed the most energy-efficient model given its compactness (22\% more energy-efficient than ours). However, when \textbf{\textit{Dyn}} optimizations are applied, our model's efficiency improves drastically to reach the \textit{same} level of energy efficiency as \textbf{a0}. Even more so, our model becomes 19\% more energy-efficient than \textbf{a0} once \textbf{\textit{Dyn w/ HW}} optimizations are in place.

\textbf{Analysis Summary and Conclusions:} Through its awareness of the \textit{dynamic} and \textit{DVFS} parameter spaces, HADAS can balance the accuracy-efficiency trade-offs more than the conventional NN design approaches. 
Specifically, HADAS's joint optimization approach of the backbone model, early exiting features, and the hardware settings leads to DyNN model designs that are highly prone to benefit from the static, dynamic, and hardware deployment aspects altogether.

% \blue{We remark that for this analysis, dynamic evaluations are assumed under ideal settings for every model, that is, input samples are mapped to the most computationally efficient exit at which it is classified correctly. we detail the design in later sections} 

% \begin{figure}[ht]
%   \centering
%       \includegraphics[width=0.45\textwidth, height=0.15\textheight]{figures/accuracy_motiv.pdf}
%       \caption{Comparison between conventional and DyNNs with regards to their TOP-1 accuracy when optimizing with early-exit and DVFS. (\textbf{a0-6} for optimized baselines, \textbf{b1-2} for optimal models obtained by HADAS)}
%       \label{fig:motiv_acc}
% \end{figure}

\vspace{-1ex}
\subsection{Novel Contributions} 
\begin{figure}[t]
   \centering
      \includegraphics[width=0.48\textwidth]{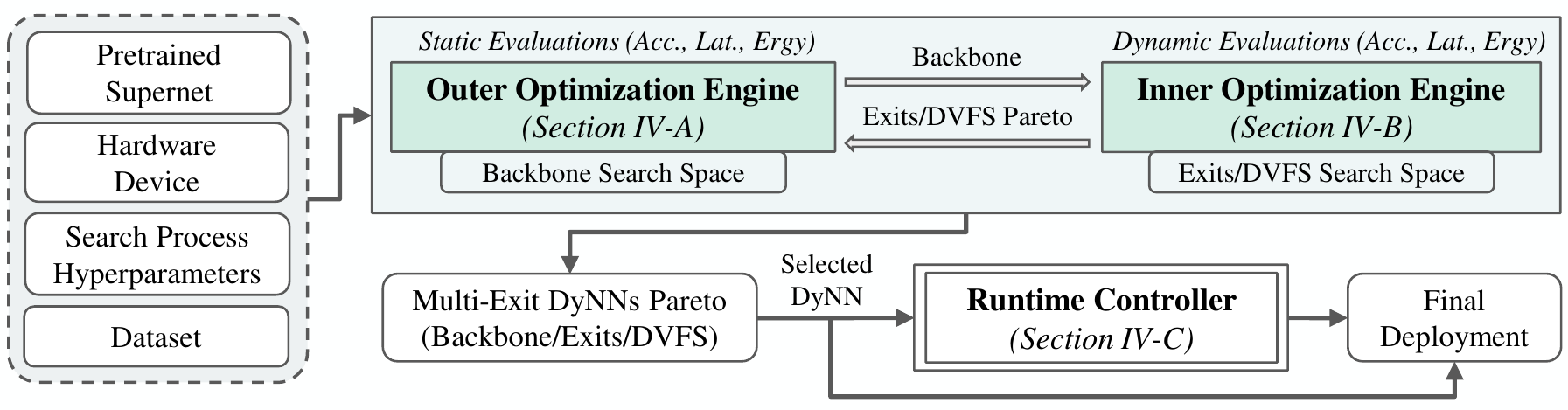}
    %   \caption{Overview of our proposed Hardware-Aware Dynamic Neural Architecture Search (HADAS) framework. Lightly-shaded green blocks represent the contributions of HADAS.}
      \vspace{-1ex}
      \caption{Overview of our Hardware-Aware Dynamic Neural Architecture Search (HADAS) framework. Lightly-shaded green blocks are the novel contributions.}
      \label{fig:HADAS_overview}
      \vspace{-3ex}
   \end{figure}

 Our scientific contributions and novelties are as follows:
\begin{enumerate}
    \item We present \textsc{HADAS}, a novel hardware-aware NAS framework that jointly optimizes the design of multi-exit DyNNs and DVFS settings for efficient edge operation.
    % \item We propose HADAS, a co-optimization framework for designing energy-efficient dynamic neural networks through the early exit and voltage/frequency scaling (DVFS).
    \item As shown in Figure \ref{fig:HADAS_overview}, \textsc{HADAS} is built to leverage the existing infrastructure of pretrained supernets provided through state-of-the-art NAS frameworks, and is also compatible with existing runtime controllers for an effective end-to-end design workflow.
    \item We formulate the design space exploration problem for multi-exit architectures as a bi-level optimization problem solved through two nested evolutionary genetic engines. The outer engine identifies optimal backbone designs. Whereas the inner engine co-optimizes the exits' integration and the DVFS settings. 
    % \item To effectively explore the joint search spaces of neural networks, early exit, and DVFS, the search algorithm of HADAS is decomposed into two evolutionary-based optimization engines. The first is an outer engine that optimizes the static inference by searching for the best backbone neural network. The second is an inner engine that aims to optimize the dynamic inference by exploring the best exit schemes and DVFS policies.
    % OD: This detail maybe for the methodology section
    % \item To keep both optimization engines updated, we propose a communication scheme in which the Pareto optimal sets are shared at the end of each optimization cycle. The outer-to-inner feedback is used to start the inner optimization engine for specific backbone neural networks. In contrast, the inner-to-outer feedback is used as an elite selection strategy for the outer optimization engine.
    \item On the CIFAR-100 dataset and a diverse set of hardware devices/settings, our experiments demonstrated that HADAS models can realize energy efficiency gains by up to \textbf{$\sim$57\%} over models designed through conventional methods while preserving the desired level of accuracy.
    %\red{Should we put here the best results on one hardware setting or different settings?} \blue{you can average or pick your best -- details of results come later}
\end{enumerate}

% --> Give an overview on dynamic neural architectures in general and on early-exit in particular

\section{Related Works}
\label{sec:related}

\textbf{Early exiting and NAS:} Early-exiting has been widely adopted to realize DyNNs on the edge given their ``simple-yet-effective'' characteristic. The direct approach to realize Multi-exit networks has been to \textit{branch} intermediate classifiers from the earlier stages of a backbone model, and retraining the model to maximize the performance of all classifiers \cite{panda2016conditional, teerapittayanon2016branchynet, Phuong_2019_ICCV, huang2017multi}. With an effective input-to-exit mapping policy, Multi-exit models enjoy computational efficiency as \textit{simpler} input samples can be classified at the earlier classifiers (exits) while maintaining the model's representational power through retaining the full classifier for the \textit{harder} samples. In the aforementioned works, the multi-exit networks have been manually designed based on heuristic choices of positions, structure, and count conditioned on their respective backbone architecture \cite{laskaridis2020hapi}. Recent works \cite{odema2021eexnas, yuan2020s2dnas} have investigated the applicability of NAS techniques to automate the design of multi-exit networks, where the backbone and exits' design spaces can be jointly explored to reach superior DyNN architectures. However, \cite{odema2021eexnas} instituted a small search space of one exit branch at a fixed position which is not scalable. Whereas despite the effectiveness of the approach in \cite{yuan2020s2dnas}, its application was specific to convolutional NNs.

\textbf{Dynamic hardware reconfiguration:}
Dynamically scaling NNs results in different computational and energy footprints that require adapting the hardware configuration accordingly. In \cite{paul2019hardware, farhadi2019novel}, the hardware has been co-designed with the multi-exit networks using FPGAs, showcasing how further energy efficiency gains can be achieved through having specialized hardware for exits. Nevertheless, the considerable switching overheads of hardware configurations in FPGAs are not typically acceptable for runtime applications. A viable alternative came in the form of hardware reconfiguration through supported DVFS features, where the operational frequency can be scaled after exiting to preserve energy resources \cite{tambe2021edgebert, li2022predictive}. 
%For instance in \cite{tambe2021edgebert}, the entropy estimates are used to scale the frequency after exiting \blue{last sentence not clear}. In Predictive Exit, \cite{li2022predictive}, the frequency is scaled to a middle value during the inference, then a low frequency is set after the exit to save energy. 
Table \ref{tab:sota_table} illustrates the difference between HADAS and existing multi-exit network design approaches and how it improves upon them through its joint optimization approach while being compatible with existing state-of-the-art NAS frameworks.
\vspace{-2ex}

\begin{table}[ht!]
\centering
\caption{Comparison between Related-works and ours 
\vspace{-1.5ex}
} 
\fontsize{9}{9}\selectfont
\scalebox{0.85}{
\label{tab:sota_table}
\begin{tabular}{ccccc} 
\hline
Work                                                             & Early-Exiting        &  NAS        & DVFS       & Compatibility  \\ 
\hline
BranchyNet~\cite{teerapittayanon2016branchynet} & x                                                                &            &            &                \\
CDLN~\cite{panda2016conditional}                & x                                                              &            &            &                \\
S2dnas~\cite{yuan2020s2dnas}                    & x                                                              & x          &            &                \\
Dynamic-OFA~\cite{lou2021dynamic}               &                                                         & x          &            & x              \\
EExNAS~\cite{odema2021eexnas}                   & x                                                             & x          &            &                \\
Edgebert~\cite{tambe2021edgebert}               & x                                                        &            & x          &                \\
Predictive Exit~\cite{li2022predictive}         & x                                                             &            & x          &                \\ 
\hline
\textbf{HADAS}                                                   & \textbf{x}                                                         & \textbf{x} & \textbf{x} & \textbf{x}              \\
\hline
\end{tabular}
}
\vspace{-3ex}
\end{table}

\section{Problem Formulation}

As the combined design space size for the DyNNs and hardware configurations can be enormous, we characterize three separate subspaces to manage the joint optimization of their parameters as follows: (\emph{i})  
\textbf{\textit{The backbones} ($\mathcal{B}$)}; which are models originally designed in a monolithic fashion for \textit{static} inference with no adaptive behavior, (\emph{ii}) \textbf{\textit{The exits} ($\mathcal{X}$)}; which are the dynamic components to be integrated onto a backbone, and (\emph{iii}) \textbf{\textit{The DVFS settings} ($\mathcal{F}$)}; constituting the space of operational frequencies for the underlying hardware components. For the DyNNs, our reasons for designating $\mathcal{B}$ and $\mathcal{X}$ as separate subspaces are twofold: (\emph{a}) To maintain the generality of the approach by having the $\mathcal{X}$ subspace indifferent to the ``\textit{type}'' of candidate backbones in $\mathcal{B}$, and (\emph{b}) To leverage the existing infrastructure of pretrained supernets from established NAS frameworks (as in \cite{wang2020attentivenas, cai2019once}) so as to provide high-caliber backbone models for the $\mathcal{B}$ subspace. 

In order to rank candidate dynamic architectural designs, we denote \textbf{$\mathcal{S}$} and \textbf{$\mathcal{D}$} as generic performance objectives under \textit{static} and \textit{dynamic} deployments, respectively. Mainly, \textbf{$\mathcal{S}$} represents the backbone evaluations when designated as a fixed standalone model (e.g., baseline energy), whereas \textbf{$\mathcal{D}$} is for the evaluations of its dynamic variant after integrating the exits (e.g., average energy when effective mapping of inputs to exits). Hence, this implies a bi-level optimization problem with the $\mathcal{B}$ as the outer-level subspace and ($\mathcal{X}$, $\mathcal{F}$) as the inner-level ones:  
\begin{align}
    &b^{*} = \argmax_{b \in \mathcal{B}}\; \psi[\mathcal{S}(b), \mathcal{D}(x^*, f^*\; |\; b)] \label{eqn:outer}\\
    &s.t.\  x^*, f^* = \argmax_{x \in \mathcal{X}, f \in \mathcal{F}} \; \mathcal{D}(x, f\; |\; b) \label{eqn:inner}
\end{align}
% \textcolor{purple}{Here I see two problems. A: arg max makes sens only when 1 function is used (ex arg max (f)), B: in the number of arguments of \mathcal{D}(...). 2 or 3?. I think we have to rewrite these two equations} 
where the global optimization objective to identify the ideal parameter combination ($b^{*}$, $x^{*}$, $f^{*}$) that maximizes a global function $\psi$ combining the performance objectives of $\mathcal{S}$ and $\mathcal{D}$. In practice, the underlying optimization objectives are conflicting by nature -- e.g., the larger computationally expensive models enjoy higher accuracy scores and vice versa. Thus, the problem can be approached as a multi-objective optimization where we seek a Pareto optimal set of solutions 
%that can dominate all others. 
For instance, in equation (\ref{eqn:inner}), a solution ($x^*$, $f^*$) is said to be Pareto optimal if for all the objective functions $d$ $\in$ $\mathcal{D}$: 
\vspace{-1ex}
\begin{align*}
    d_{k}(x^{*}, f^{*}) \geq &d_{k}(x, f) \forall k,(x,f) \ \\ &\text{and} \ \exists j: d_{j}(x^{*}, f^{*}) > d_{j}(x,f) \forall (x, f) \neq (x^{*}, f^{*}) 
\end{align*}

\begin{figure}[t!]
\centering
\includegraphics[width=0.48\textwidth]{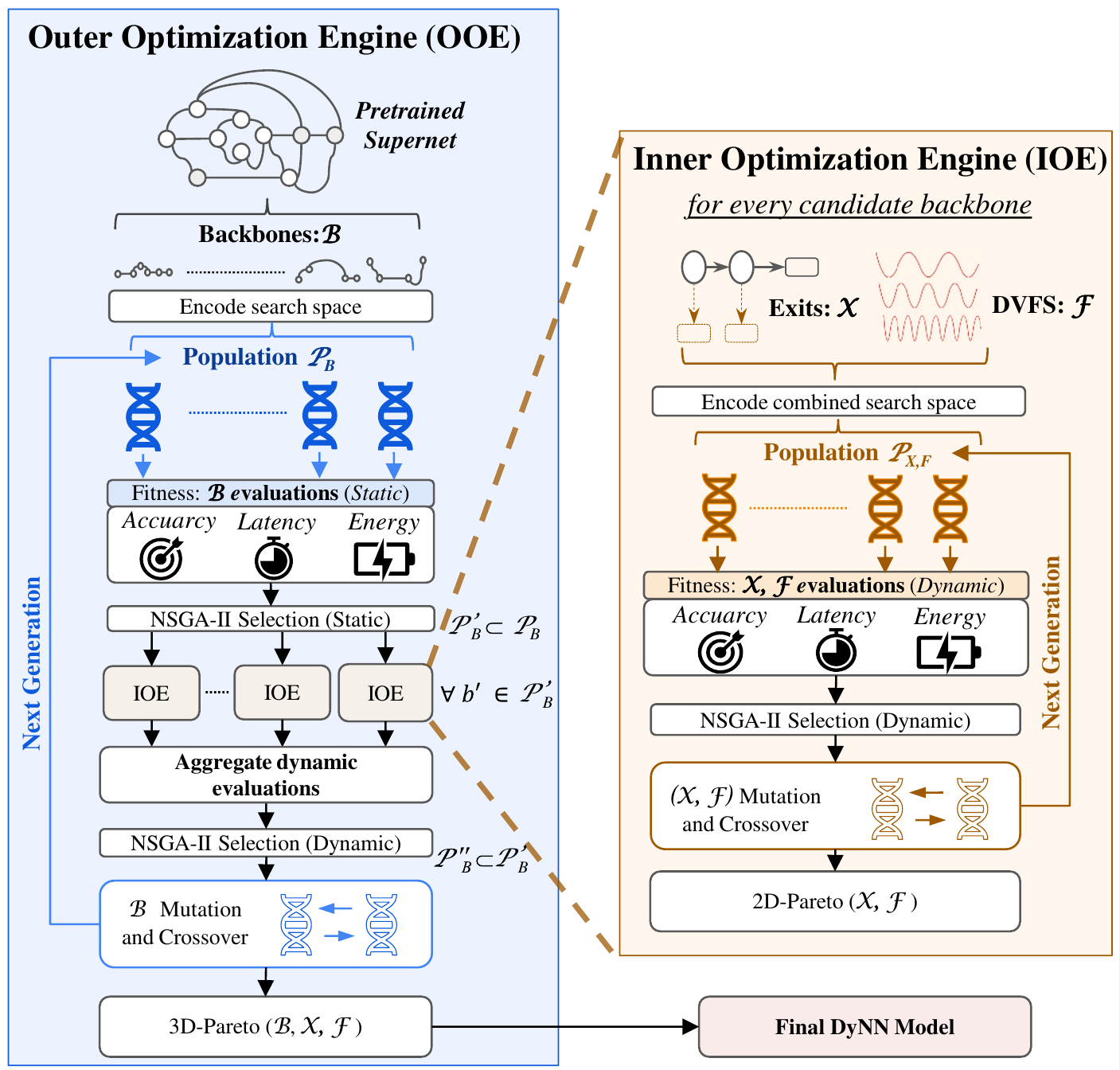} 
\caption{HADAS co-optimization framework.}
\label{fig:approach}
\vspace{-3ex}
\end{figure}

\section{Hadas Framework} \label{sec:approach}

We adopt nested genetic algorithms \cite{fasfous2022anaconga} to solve the optimization problem as illustrated in Figure \ref{fig:approach} as follows:

% illustrates HADAS's nested outer and inner optimization engines -- OOE and IOE -- detailed as follows:

% Briefly, the design subspaces are defined prior to the initiation of the optimization process where: $\mathcal{B}$ encompasses the set of pretrained backbones in a NAS supernet, $\mathcal{X}$ covers the count and position space of early-exits, and $\mathcal{F}$ indicates the configurable DVFS frequencies on the target hardware. The details are provided in the following subsections. 

% inputs to the framework include the search space definition, the hardware device, the Early-exit search space (i.e., number and positions), and the available parameters of frequencies on the hardware device. HADAS optimizes three search spaces through two optimization engines: 1) An outer optimization engine that optimizes the neural network architectures, and 2) An inner optimization engine that searches for the best early-exit strategy and DVFS policy for the dynamic inference. The communication between the two engines is ensured by employing feedback from the optimization cycles of each optimization engine. \blue{we should mention the timing overheads of the inner and outer}

\subsection{Outer Optimization Engine (OOE)}

The OOE considers two primary tasks: (\emph{i}) Searching through $\mathcal{B}$ to identify the best backbone candidates, and (\emph{ii}) Ranking DyNNs according to their aggregate $\mathcal{S}$ and $\mathcal{D}$ evaluations. 

% We break down the OOE following the left part of Figure \ref{fig:approach}:

\subsubsection{$\mathcal{B}$ Subspace} Modern NAS frameworks employ a Once-For-All (OFA) approach which entails first training a large over-parameterized \textit{supernet} on a target task, prior to applying a search algorithm to identify the optimal subnet designs within. The enabling factor of OFA approaches is that all of the supernet's parameters are \textit{shared} by its subnets, effectively rendering the \textit{training} and \textit{search} procedures as disjoint processes, which dramatically reduces the overall overheads within the NAS framework \cite{cai2019once, wang2020attentivenas}. From here, HADAS is built to leverage the pretrained supernets of existing NAS frameworks to construct the $\mathcal{B}$ subspace of backbones, where the search space can be \textit{encoded} into discrete variables usable by the search algorithm, and each viable subnet (backbone) can be denoted as $b\in \mathcal{B}$. 
% In our experiments, we leverage the AttentiveNAS supernet \cite{wang2020attentivenas} that contains more than $2.94\times10^{11}$ candidate backbones, with the $\mathcal{B}$ search parameters defined in Table \ref{tab:joint_search_space}.

% We use a Once-For-All neural network for the NAS process. Specifically, our OFA neural network is based on the AttentiveNAS \cite{wang2020attentivenas} pretrained supernet that contains more than $2.94\times10^{11}$ neural networks generated by scaling different dimensions, including the input resolution, depth, width, kernel size, and expand ratio. Thus, all subnets share the same parameters, allowing a dynamic switching at runtime from one subnet to another with minimal switching overhead.

\subsubsection{$\mathcal{B}$ Evolutionary Search}
With $\mathcal{B}$ defined, the dynamic architecture search initiates in the OOE through an evolutionary search algorithm (e.g., NSGA-II) that can navigate through $\mathcal{B}$ to sample promising backbone models. In particular, the evolutionary algorithm is set to run for a predefined number of generations $G$, generating with every generation, $g$, a population of backbones, $\mathcal{P}_\mathcal{B}^g$, from which the encoded pretrained subnets can be sampled. Afterwards, $\forall b \in {P}_\mathcal{B}^g$, a fitness evaluation under \textit{static} conditions is performed as: 
% \blue{@halima was this a weighted sum or multiplication - let me know i can write it that way? }
% \red{We can write it this way as we need after a vector of objectives to perform the NSGA-II selection. The vectors are evaluated separately, we don't apply any summation or multiplication on them}
\vspace{-0.5ex}
\begin{equation}
    \mathcal{S}(b) = Fit(Acc_b, L_b, E_b)
\end{equation}

where $\mathcal{S}(b)$ is a vector of the \textit{static} performance evaluations with regards to the accuracy ($Acc_b$), latency ($L_b$), and energy ($E_b$), respectively. Estimates for $L_b$ and $E_b$ are obtained based on hardware measurements -- as through a HW-in-the-loop setup (adopted here), lookup tables, or prediction models. At this stage, we remark that hardware evaluations are based on default HW settings, leaving the DVFS optimizations for the IOE. Based on the $\mathcal{S}$ scores, every $b \in {P}_{\mathcal{B}}^g$ is ranked using the NSGA-II non-dominated sorting algorithm. If a number of backbones shared the same rank, their diversity scores are used for re-ranking. This early selection procedure enables pruning the population to reach a smaller subset $\mathcal{P}_{\mathcal{B}}^{g'}$ $\subset$ $\mathcal{P}_\mathcal{B}^{g}$, where every $b' \in {P}_\mathcal{B}^{g'}$ is mapped to an IOE (detailed later) to obtain the overall dynamic  architecture evaluations $\mathcal{D}(x^{*}, f^{*}\; |\; b')$. 

Once an IOE concludes its procedures, a Pareto optimal set of exits placement and DVFS settings is returned to the OOE for every $b'\in {P}_{\mathcal{B}}^{g'}$. These Pareto sets are then collectively aggregated for a second selection algorithm that ranks backbones based on their combined $\mathcal{S}$ and $\mathcal{D}$ scores, leading to another population subset $\mathcal{P}_{\mathcal{B}}^{g''} \subset \mathcal{P}_{\mathcal{B}}^{g'}$. Lastly, $\mathcal{P}_{\mathcal{B}}^{g''}$ undergoes \textit{mutation} and \textit{crossover} operations to construct a new population $\mathcal{P}_\mathcal{B}^{g+1}$ for generation $g$+1. This outer loop cycle repeats until generation $G$ at which the Pareto optimal set ($b^{*}$, $x^{*}$, $f^{*}$) is returned as the final solution.

\subsection{Inner Optimization Engine (IOE)}
The IOE is invoked for every $b' \in \mathcal{P}_{\mathcal{B}}^{g'}$. Its primary responsibility is to search through the defined $\mathcal{X}$ and $\mathcal{F}$ subspaces to identify optimal pairings ($x^{*}$, $f^{*}\; | \; b'$) as follows:

\subsubsection{$\mathcal{X}$ subspace}
To define the exits' search space, we characterize the total \textit{number} of exits and their \textit{positions} as search parameters. In practice, present-day backbone structures (as those from AttentiveNAS) constitute $M$ sequential computing neural blocks (i.e., an aggregation of interrelated layers) between which effective placement of the exits can be realized. We illustrate this in Figure \ref{fig:exit_arch} through how the $\mathcal{X}$ subspace is conditioned on a $b \in \mathcal{B}$. Specifically, we define a vector of indicators $[\mathcal{I}_{1}, \mathcal{I}_{2}, ..., \mathcal{I}_{M-1}]$ where $\mathcal{I}_i \in \{0,1\}$ to indicate whether exit branch at position $i$ is sampled for the corresponding instance. Regarding the composition of exit branches, we fix a simple structure across all potential exits positions for three reasons: (\emph{i}) Re-usability as such a straightforward structure can act as a base module compatible with numerous backbone model architectures and classes, (\emph{ii}) The smaller search space size of the exits leads to smaller search overheads -- especially relevant when considering the additional subspaces as well, and (\emph{iii}) Minimizing the training costs of the exits. For our experiments, the exit structure constituted a single sequential computing block of a convolutional, batch normalization and activation layers, which are followed by a final classifier layer.

\subsubsection{Exits Training} 
Once a $b'$ is mapped to the IOE, every $x \in \mathcal{X}$ needs to be trained for a fair evaluation of the exit candidates. In this scheme, the weight parameters of $b'$ are kept \textit{frozen} independent of the exits' training procedure, where the rationale here is to avoid negatively influencing the performance of $b'$ with regards to its static accuracy score (i.e., the backbone accuracy) -- which can occur when the weights are optimized for more than one objective \cite{teerapittayanon2016branchynet}. Combining this notion with the compact structure of the exits, the exits' training overheads can be kept to a minimum within the IOE, all while leveraging the representational power of $b'$ across its various stages to attain the desired resource efficiency gains. 

\begin{figure}[t]
   \centering
      \includegraphics[width=0.48\textwidth]{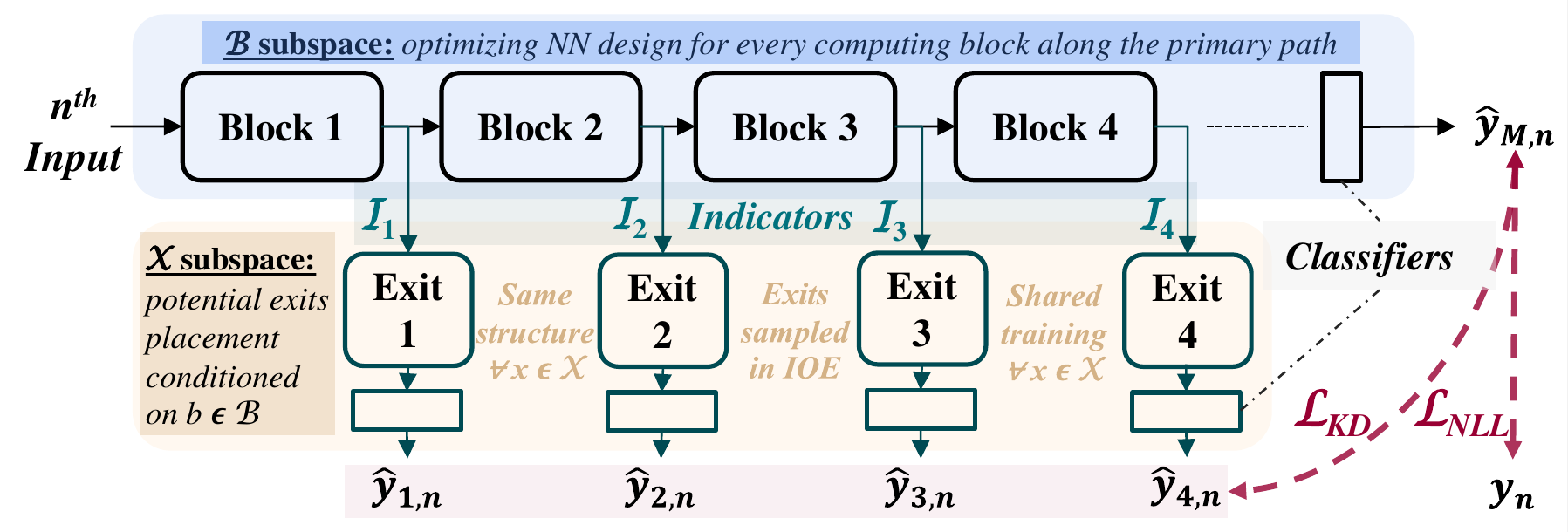}
      \vspace{-1.5ex}
       \caption{The combined $\mathcal{B}$ and $\mathcal{X}$ search spaces}
      \label{fig:exit_arch}
      \vspace{-3.5ex}
   \end{figure}

For the training loss function itself, we adopt a hybrid loss function ($\mathcal{L}_{total}$) combining the Negative log-likelihood ($\mathcal{L}_{NLL}$) and knowledge distillation ($\mathcal{L}_{KD}$) loss components to simultaneously train every $x \in \mathcal{X}$ as follows:
\vspace{-0.8ex}
\begin{equation}
    \mathcal{L} = \frac{1}{N}\sum_{n=1}^{N} [\frac{1}{M\text{-1}}\sum_{m=1}^{M-1}(\mathcal{L}_{NLL}(y_{n}, \hat{y}_{m,n}) + \mathcal{L}_{KD}(\hat{y}_{m,n}, \hat{y}_{M,n})]
\label{eq:exit_loss}
\end{equation}
% \begin{equation}
%     \mathcal{L}oss = \frac{1}{N}\sum_{n=1}^{N} [\frac{1}{M}\sum_{m=1}^{M}(NLL(x_{m,n}, y_{m,n}) + KD(x_{m,n}, x_{M,n}))]
% \label{eq:exit_loss}
% \end{equation}
where $N$ is the total number of training samples and $M-1$ is the total possible number of exits. For the $\mathcal{L}_{NLL}$ term, it aggregates the losses from every exit at $m$ when comparing its predicted outputs, $\hat{y}_{m,n}$, against the ground truth labels, $y_{n}$, for every sample $n$. Whereas the $\mathcal{L}_{KD}$ term aggregates the losses from comparing the error between every $\hat{y}_{m,n}$ and that of the final model classifier, $\hat{y}_{M,n}$. Due to space limitations, we illustrate how these loss components are defined in Figure \ref{fig:exit_arch}, and refer interested readers to \cite{Phuong_2019_ICCV} for more details.
% \begin{equation}
%     loss = \frac{1}{N}\sum_{n=1}^{N} [\frac{1}{M}(nll(x_{m,n}, y_{m,n}) + dist(x_{m,n}, x_{M,n}))]
% \label{eq:exit_loss}
% \end{equation}

\subsubsection{$\mathcal{F}$ subspace} The hardware search space entails the DVFS configurations for enhancing the DyNN's resource efficiency from the HW's perspective. Given how different computational workloads utilize the underlying hardware components differently, DyNN design candidates can attain maximal resource efficiency at different DVFS settings. In practice, edge devices constitute heterogeneous computing units that support DVFS features. Thus, depending on the underlying hardware, the operational frequencies of CPU, GPU, and External Memory Controllers (EMC) can be used to construct $\mathcal{F}$.

\subsubsection{($\mathcal{X}$, $\mathcal{F}$) Evolutionary Search}
Similar to the OOE, an IOE also operates an evolutionary NSGA-II algorithm to navigate the combined search spaces of $\mathcal{X}$ and $\mathcal{F}$. With each generation, a population $\mathcal{P}_{\mathcal{X}, \mathcal{F}}$ is generated from the combined subspaces' encoding and provided for the dynamic fitness evaluation:
\vspace{-1ex}
\begin{align}
    &\mathcal{D}(x, f\; |\; b') = \frac{1}{\sum_{i=1}^{M\text{-1}} \mathcal{I}_i} \;\sum_{i=1}^{M\text{-1}}\; \mathcal{I}_i \cdot [score_i] \label{eq:inner_objective} \\
    &s.t.\;\; score_i = \mathcal{N}_i * \frac{E_{x_i,f}}{E_{b}} * \frac{L_{x_i,f}}{L_{b}} * (dissim_i)^{\gamma} \label{eq:exit_score}
\end{align}
where equation (\ref{eq:inner_objective}) reflects the mean dynamic performance score of a sampled dynamic model ($x$, $f\; |\; b'$) through averaging scores for every sampled exit (recall $\mathcal{I}_i \in \{0,1\}$). An exit's score is given by $score_i$ in equation \ref{eq:exit_score}, which constitutes: $\mathcal{N}_i$, the fraction of samples that can be correctly classified at exit $i$; $\frac{E_{x_i,f}}{E_{b}}$, as the normalized dynamic energy at exit $x_i$ and DVFS settings $f$ relative to the backbone energy consumption; $\frac{L_{x_i,f}}{L_{b}}$ is similarly the normalized dynamic latency term. $(dissim_i)^{\gamma}$ is a regularization term with a trade-off parameter $\gamma$ measuring the dissimilarity of exit $x_i$ and its preceding ones as:
\begin{equation}
    dissim_i = 1-\max\; (\mathcal{N}_{0:i-1})
\end{equation}
where $x_i$'s score is regularized in proportion to the fraction of samples that can be already classified by its preceding exits. The rationale behind this metric is to: (\emph{i}) avoid sampling exits of similar performance characterizations, and (\emph{ii}) realize a compact decision space for the DyNN when deployed. 

Based on the $\mathcal{D}$ scores, every $(x, f\; |\; b') \in {P}_{\mathcal{X}, \mathcal{F}}$ is also ranked using the NSGA-II non-dominated sorting algorithm so as to realize subset ${P}_{\mathcal{X}, \mathcal{F}}' \subset {P}_{\mathcal{X}, \mathcal{F}}$ that would then undergo \textit{mutation} and \textit{crossover} for the following generation. This loop cycle continues until the final generation where a 2-D Pareto optimal set ($x^{*}$, $f^{*}\; | \; b'$) is returned to resume the OOE.

\subsection{Runtime Controller}

When a DyNN design is chosen for the final deployment, a runtime controller needs to be implemented to provide the effective input-to-exit mapping policies needed for dynamic inference. Concerning \textsc{HADAS}, its architectural optimizations are applied at the design stage of DyNNs under \textit{ideal} mapping policies, that is, when every input is mapped to the first exit module $x_i$ that can classify it correctly. This is evident through how the score of each exit in eq. (\ref{eq:exit_score}) is scaled based on $\mathcal{N}_i$ -- the \textit{true} fraction of correctly classified samples. Thus, models from \textsc{HADAS} are compatible with any class of runtime controllers existing in the literature (e.g., entropy-based \cite{panda2016conditional, teerapittayanon2016branchynet, han2021dynamic}).

\section{Experiments and Results}
\subsection{Experimental Setup} 

% Joint search space, hardware settings, and CIFAR-100 
We implement HADAS on top of the AttentiveNAS framework \cite{wang2020attentivenas}. To construct $\mathcal{B}$, we reuse their search space which contains more than $2.94\times10^{11}$ neural networks generated by scaling different dimensions as stated in Table \ref{tab:joint_search_space}. Our experiments are conducted on the CIFAR-100 dataset where the pretrained supernet of AttentiveNAS has been fine-tuned accordingly. Backbones and baselines are all sampled from the same fine-tuned supernet. We dynamically generate the exits' search space $\mathcal{X}$ according to the supported depth $(l)$ of the backbones in $\mathcal{B}$. In our case, potential exit positions occur at a layer-wise granularity starting from the fifth (\nth{5}) layer to the backbones' last layer (For AttentiveNAS \cite{wang2020attentivenas}, potential exit positions are set after their ``MBConv'' layers). 
We evaluate our approach on 4 different hardware combinations from NVIDIA Edge devices: a) \textbf{\textit{AGX Volta GPU}}, b) \textbf{\textit{Carmel ARM v8.2 CPU}}, c) \textbf{\textit{TX2 Pascal GPU}}, and d) \textbf{\textit{NVIDIA Denver CPU}}.
For each hardware setting, we leverage the supported DVFS configuration settings to generate $\mathcal{F}$ as in Table \ref{tab:joint_search_space}. Regarding the optimization process, we fix a budget of 450 iterations for the OOE and 3500 iterations for the IOE, where \#iterations = $\mathcal{G}$ $\times$ $\mathcal{P}$. We use a cluster of 32 GPUs to train the exits for every sampled backbone, taking up to $\sim$ 8-10 GPU hours for each $\mathcal{G}$. In our experiments, we used a HW-in-the-loop setup for latency and energy measurements which pushed the overall search time of HADAS to $\sim$2-3 GPU days. Nevertheless, based on our analysis, HADAS's search overhead can be reduced to 1 GPU day if a %lookup table or a 
proxy model replaced the HW-in-the-loop setup. 

\begin{table}[t!]
\centering
\caption{Details on HADAS joint search spaces in our experiments} 
\fontsize{9}{9}\selectfont
\scalebox{0.8}{
\label{tab:joint_search_space}
\begin{tabular}{lll} 
\hline
\multicolumn{1}{l}{\textbf{Decision variables}} & \multicolumn{1}{l}{\textbf{Values}} & \multicolumn{1}{l}{\textbf{Cardinality}}  \\ 
\hline
\multicolumn{3}{c}{\textbf{Backbone Search Space ($\mathcal{B}$)}}                                                                                   \\ 
\hline
Number of blocks (n\_block)                     & 7                                   & 1                                         \\
Input resolution (res)                          & \{192, 224, 256, 288\}              & 4                                         \\
Block depth (l)                                 & \{1, 2, 3, 4, 5, 6, 7, 8\}          & 8                                         \\
Block width (w)                                 & {[}16, 1984]                        & 16                                        \\
Block kernel size (k)                           & \{3, 5\}                            & 2                                         \\
Block expand ratio (er)                         & \{1, 4, 5, 6\}                      & 4                                         \\ 
\hline
\multicolumn{3}{c}{\textbf{Exits Search Space ($\mathcal{X}$)}}                                                                                     \\ 
\hline
Number of exits (nX)                      & {[}1, ($\sum_{i=1}^{nb}l_i) - 5$]                          & max(nX)                                       \\
Exit positions (posX)                     & {[}5, $\sum_{i=1}^{nb}l_i)$]                            &  
$\binom{nx}{\sum_{i=1}^{nb}l_i)}$   \\ 
\hline
\multicolumn{3}{c}{\textbf{DVFS Search Space ($\mathcal{F}$)}}                                                                                    \\ 
\hline
GPU frequency (AGX Volta GPU)                   & {[}0.1GHz, 1.4GHz]                  & 14                                        \\
CPU frequency (Carmel ARM v8.2 CPU)                 & {[}0.1GHz, 2.3GHz]                  & 29                                        \\
GPU frequency (TX2 Pascal GPU)                  & {[}0.1GHz, 1.4GHz]                  & 13                                        \\
CPU frequency (NVIDIA Denver CPU)                  & {[}0.3GHz, 2.1GHz]                  & 12                                        \\
EMC frequency (AGX SOC)                         & {[}0.2GHz, 2.1GHz]                  & 9                                         \\
EMC frequency (TX2 SOC)                         & {[}0.2GHz, 1.8GHz]                  & 11                                        \\
\hline
\end{tabular}
}
\vspace{-4ex}
\end{table}

% However, we noticed that the main bottleneck was on the HW-in-the-loop evaluation, which takes $\sim$ 6-8 hours to run the IOE for all the sampled backbones from the OOE, resulting in a timing overhead of $\sim$ 4-5 GPU days to run HADAS on a single hardware setting.

\subsection{Co-optimization Results} 

%\subsection{Bi-level optimization with variable HW config} 
% We can divide into 2 separate subsections 
% Hypervolume

\begin{figure*}[!ht]
\begin{center}
{\includegraphics[,width = .91\textwidth]{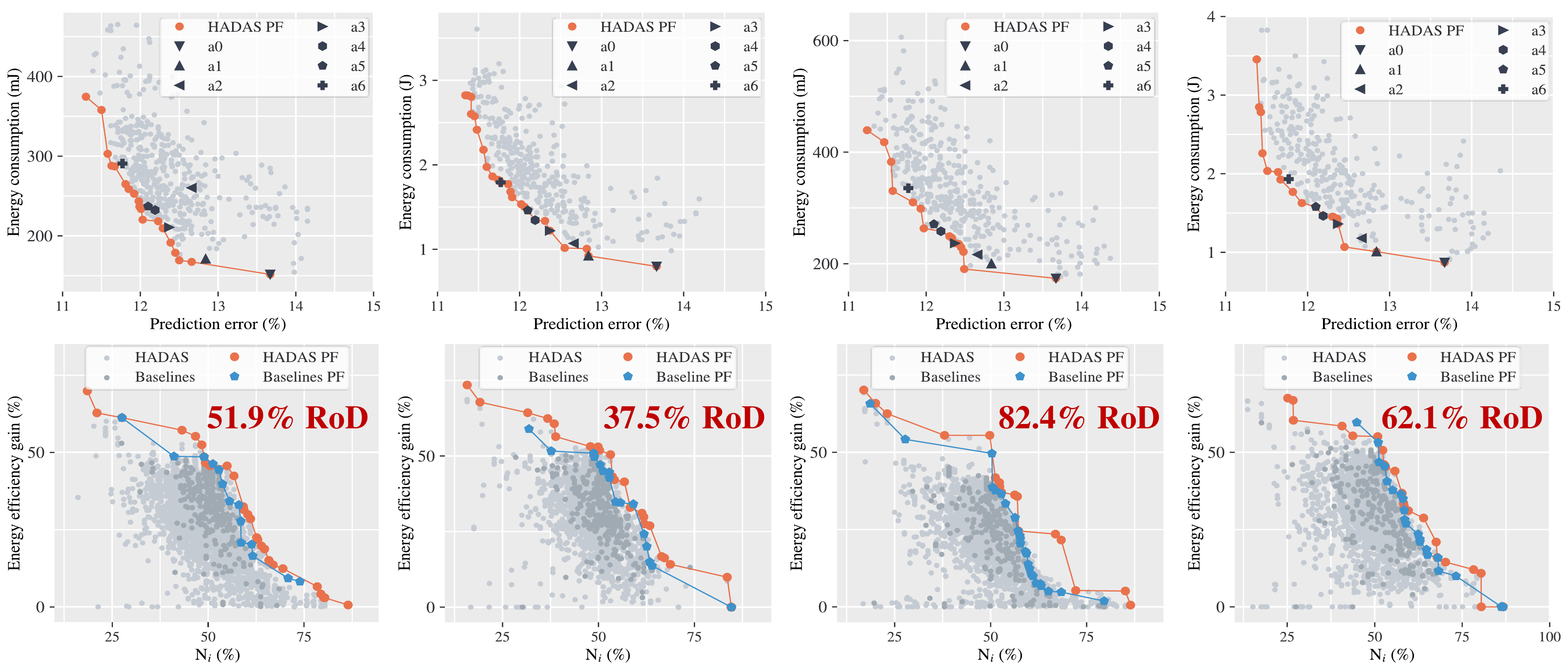}}
\end{center}
\vspace{-3ex}
\caption{The top row gives the results of the outer optimization on 4 hardware settings of (\textit{from left to right}): a) AGX Volta GPU, b) Carmel ARM v8.2 CPU, c) TX2 Pascal GPU, and d) NVIDIA Denver CPU. The bottom row shows the results of the inner optimization engine, with the same hardware settings.  
The points in the top row depict the static performance of the explored backbone neural networks in $(\mathcal{B})$ by the OOE, without early-exit or DVFS. The points in the bottom row represent the performance of the explored combinations of backbones, early-exits, and DVFS in $(\mathcal{B}, \mathcal{X}, \mathcal{F})$ by the IOE.
} 
\label{fig:co-opt-results}
\vspace{-3ex}
\end{figure*}

\textbf{OOE Analysis:}
The top row of Figure \ref{fig:co-opt-results} compares the static performance results from the OOE of HADAS against those of the top models from AttentiveNAS \cite{wang2020attentivenas} (denoted as \textbf{[a0-a6]}). As shown, our obtained Pareto fronts (PF) generally dominate the baselines on the four hardware settings. 
Furthermore, HADAS can identify comparable backbones to the baselines with just a few evaluations. 
For instance, on the AGX Volta GPU, \textbf{a6} is dominated by another backbone from HADAS with an energy reduction of $\sim33\%$ under the same accuracy level. Similarly, \textbf{a1} is dominated by another backbone from HADAS with an accuracy improvement of \textbf{2.34\%} under the same energy gain.
%(450 evaluations against 10240 evaluations in AttentiveNAS \blue{but their number is on imagenet, no?}) \red{yes, maybe this is not a fair comparison?}. 
%Overall, an accuracy improvement up to $\sim$ 0.4\% and energy reduction up to $\sim$ 25\% were achieved by HADAS over the baselines of AttentiveNAS. \blue{this is vague}

\textbf{IOE Analysis:}
The results of the IOE are shown in the bottom row of Figure \ref{fig:co-opt-results}. For a fair comparison, we fix the same optimization budget when running the IOE for the baselines and HADAS. The dynamic performance of the explored $(b, x, f)$ combinations and the obtained Pareto fronts are given for both approaches, where the dynamic comparison metrics are the energy efficiency gains when early exiting and DVFS are supported, as well as the average of $N_i$ values from equation (\ref{eq:exit_score}). Across the four hardware settings, HADAS seemingly dominates the majority of the optimized baselines with an average ratio of dominance \textbf{58.4\%} (detailed in the following paragraph). This can be attributed to HADAS's better understanding of the global search space, where it samples backbones that are more poised to benefit from the IOE optimizations with regard to early exiting and DVFS.
This is also evident through how HADAS can sample dynamic parameters for its models that can realize substantial energy or accuracy gains near the extremes of its Pareto frontier, which are not realizable by the optimized baselines. For instance on the Caramel ARM v8.2 CPU, energy gains reach \textbf{63\%} for one of the extreme dynamic models on the Pareto frontier of HADAS, compared to \textbf{52\%} for the extreme dynamic variant from the optimized baselines, under the same level of accuracy.

% On the one hand, by enlarging the IOE search space with early-exit and DVFS, the search algorithm reduce the energy consumption by up to $\sim$ 78\% on different hardware settings \blue{of what? one model or the average of models} \red{from the point with the highest energy gain, the extreme left on each hardware setting}. Moreover, under a high accuracy of exits (usually, when choosing the last layers as exits), the energy gains can be up to $\sim$ 32\% from adapting the DVFS \blue{i'm looking at the figures and i do not find these numbers?} \red{The middle of the Pareto front, should I add them?}. 

\begin{figure}[t!]
   \centering
      \includegraphics[width=0.48\textwidth]{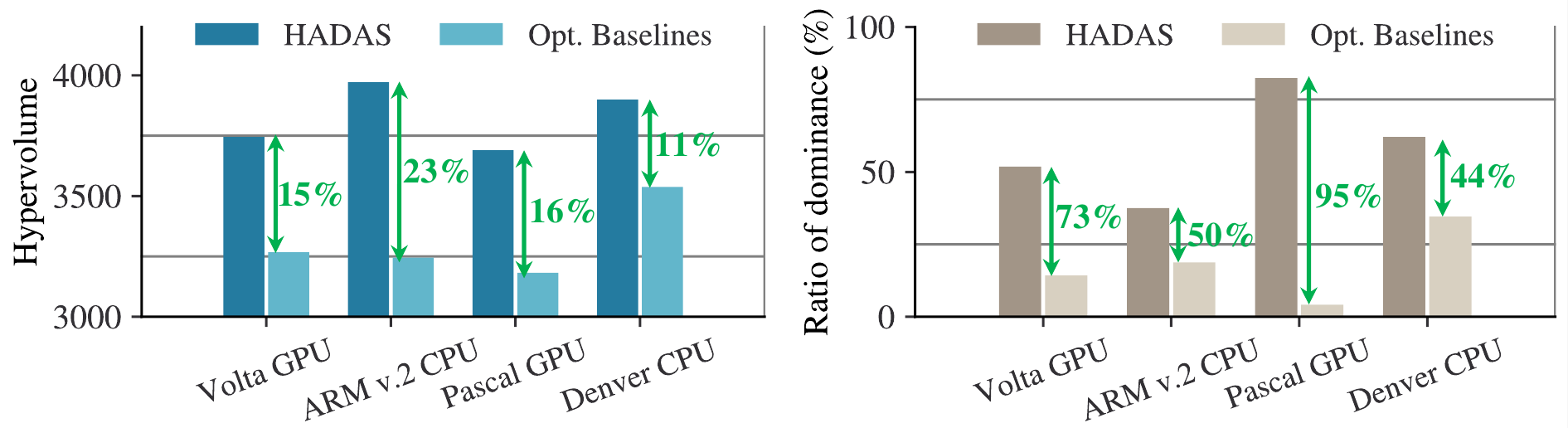}
      \vspace{-2ex}
      \caption{Comparing search efficacy for HADAS and the optimized baselines with regards to: a) hypervolume (\textit{left}) and b) ratio of dominance (\textit{right})}
      \label{fig:opt_metrics}
      \vspace{-3.3ex}
\end{figure}

% \noindent
\textbf{Hypervolume (HV) and Ratio of Dominance (RoD):}
we expand further on the IOE analysis and leverage \textit{hypervolume (HV)} and \textit{ratio of dominance (RoD)} as comparative evaluation metrics. The former metric measures the volume of the dominated portion of the objective space, whereas the latter measures the percentage of solutions found by HADAS that dominate the optimized baselines (and vice-versa). 
Figure \ref{fig:opt_metrics} shows that HADAS consistently outperforms the optimized baselines with regards to both metrics across the 4 hardware platforms. Taking the Pascal GPU as an example, we find that the \textit{HV} coverage and \textit{RoD} are \textbf{16\%} and \textbf{95\%} more for HADAS over the optimized baselines, respectively.

% \noindent
\vspace{-2ex}
\begin{table}[ht!]
\centering
\caption{DyNNs Comparison using the TX2 Pascal GPU} 
\vspace{-0.5ex}
\fontsize{9}{9}\selectfont
\scalebox{0.8}{
\label{tab:compa_sota}
\begin{tabular}{cccccc} 
\hline
\textbf{Model}   & \begin{tabular}[c]{@{}c@{}}\textbf{Baseline }\\\textbf{Acc(\%)}\end{tabular} & \begin{tabular}[c]{@{}c@{}}\textbf{EEx}\\\textbf{Acc(\%)}\end{tabular} & \begin{tabular}[c]{@{}c@{}}\textbf{Baseline }\\\textbf{Ergy(mJ)}\end{tabular} & \begin{tabular}[c]{@{}c@{}}\textbf{EEx }\\\textbf{Ergy(mJ)}\end{tabular} & \begin{tabular}[c]{@{}c@{}}\textbf{EEx\_DVFS }\\\textbf{Ergy(mJ)}\end{tabular}  \\ 
\hline
AttentiveNAS\_a0 & 86.33                                                                         & 89.95                                                                          & \textbf{173.78}                                                                 & \textbf{119.83}                                                            & \textbf{116.14}                                                                    \\
AttentiveNAS\_a6 & \textbf{88.23}                                                                & \textbf{93.02}                                                                 & 335.48                                                                          & 256.80                                                                     & 218.34                                                                             \\ 
\hline
HADAS\_b1        & \textbf{87.34}                                                                & \textbf{93.16}                                                                 & \textbf{212.44}                                                                 & \textbf{119.84}                                                            & \textbf{93.78}                                                                     \\
HADAS\_b2        & 88.06                                                                & 91.83                                                                 & 341.3                                                                           & 187.92                                                                     & 126.06                                                                             \\
HADAS\_b3        & 86.54                                                                         & 88.31                                                                          & 205.48                                                                & 130.20                                                            & 86.84                                                                     \\
HADAS\_b4        & \textbf{88.40}                                                                & 89.24                                                                          & 358.01                                                                          & 232.77                                                                     & 201.01                                                                             \\
\hline
\end{tabular}
}
\end{table}

\vspace{-1ex}

\textbf{DyNNs comparison:}
In Table \ref{tab:compa_sota}, we compare the top DyNNs obtained by HADAS with two AttentiveNAS models: \textbf{a0}, the most energy-efficient baseline, and \textbf{a6}, the most accurate baseline. Models are compared with regards to their \textit{static} (i.e., baseline accuracy and energy) and their \textit{dynamic} performances (i.e., accuracy and energy with early exiting and DVFS). As shown, the optimal models from HADAS outperform the baselines of AttentiveNAS in both \textit{static} and \textit{dynamic} evaluations. For instance, \textbf{b1} from HADAS is \textbf{57\%} and \textbf{19\%} more energy-efficient than the \textbf{a6} and \textbf{a0}, respectively, while enjoying similar accuracy scores like the most accurate model \textbf{a6}. 

\subsection{Dissimilarity Ablation Study}
\begin{figure}[t!]
   \centering
      \includegraphics[width=0.45\textwidth]{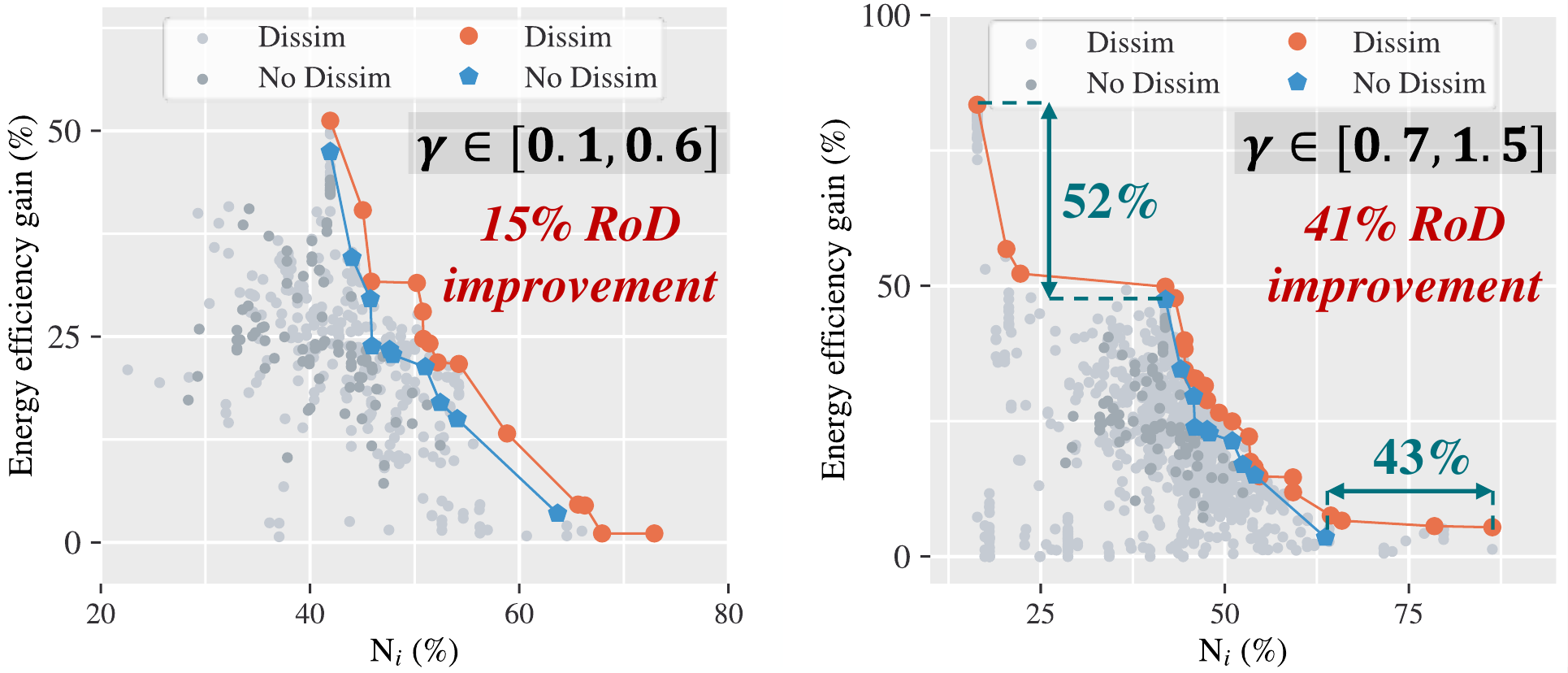}
      \vspace{-2ex}
      \caption{Inner optimization improvement by regularizing the exits scores with the dissimilarity function $(dissim)^{\gamma}$ over two ranges of $\gamma$ values}
      \label{fig:dissim_ablation}
      \vspace{-4ex}
\end{figure}

We perform an ablation study to investigate the impact of the dissimilarity term ($dissim^{\gamma}$) in equation (\ref{eq:exit_score}) through the performance of the explored models under each case. Specifically, we run the IOE for one backbone twice, with $dissim^{\gamma}$ not included and one when it is included. In Figure \ref{fig:dissim_ablation}, we compare the results obtained with and without the dissimilarity with different values of $\gamma$. As shown, the inclusion of the dissimilarity term allows the optimization algorithm to focus more on exploring dissimilar early exits with a high contribution to the prediction accuracy. 
%under different values of $\gamma$. 
For instance, in the right of Figure \ref{fig:dissim_ablation}, we find that the inclusion of dissimilarity improves \textit{RoD} by 41\%. Moreover, the extreme Pareto models with dissimilarity are $\sim$ 43\% and $\sim$ 52\% more accurate and energy efficient than those without dissimilarity.

% \vspace{-5ex}

% \vspace{-3ex}

% \vspace{-2.5ex} 

% \blue{can we replace the following lines with numbers from the figure}Hence, DyNNs with fewer dissimilar exits will be more selected by the IOE than DyNNs with many similar exits. From figure \ref{fig:dissim_ablation}, we can observe that the IOE with dissimilarity regularization can explore more accurate and energy-efficient DyNNs, especially with high values of $\gamma$. 

%\vspace{-3ex}

%\vspace{-3ex}

\vspace{-1ex}
\section{Conclusion}
We have presented \textsc{HADAS}, a novel HW-aware NAS framework that jointly optimizes the backbone, early exiting features, and DVFS for DyNNs. Through HADAS, large agile models can be realized with similar energy efficiency to that of compact models. 
We have shown that \textsc{HADAS} DyNNs can achieve up to 57\% energy gains while retaining desired accuracy levels. 
\vspace{-1ex}

% Future research can investigate how HADAS would perform on other datasets and classes of neural architectures (e.g., Transformers, Graph Neural Networks) as well as explore how to tune the DyNN to maintain optimality if the hardware settings change after deployment. 

% \blue{can mention that this design analysis is performed under ideal input-to-exit mappings. Future research can still explore the gap existing between the runtime controllers' performance and the ideal mappings}

\bibliographystyle{IEEEtran}
\bibliography{references_short}

\end{document}